\documentclass[11pt,a4paper]{article}
\usepackage[hyperref]{acl2020}
\usepackage{times}
\usepackage{latexsym}
\usepackage{amssymb}
\usepackage{mfirstuc}
\newcommand{\ra}{\(\shortrightarrow\)}
\newcommand{\type}[1]{\textsc{\capitalisewords{\MakeLowercase{#1}}}}
\newcommand{\typetwo}[2]{\type{#1}\ra \type{#2}}
\usepackage[normalem]{ulem}

\usepackage{microtype}
\usepackage{stmaryrd}
\aclfinalcopy 

\usepackage{csquotes}

\title{SERRANT: a syntactic classifier for English Grammatical Error Types}

\author{ 
  Leshem Choshen\\
  Department of Computer Science\\
  Hebrew University of Jerusalem\\
  {\tt\small leshem.choshen@mail.huji.ac.il}\\
  \And
  Matanel Oren\\
  Department of Computer Science\\
  Hebrew University of Jerusalem\\
  {\tt\small matanel.oren@mail.huji.ac.il }\\
  \AND 
  Dmitry Nikolaev\\
  Department of Linguistics \\
  Stockholm University \\
  {\tt\small dnikolaev@fastmail.com}\\
  \And
  Omri Abend \\
  Department of Computer Science\\
  Hebrew University of Jerusalem\\
  {\tt\small omri.abend@mail.huji.ac.il}\\
}
\date{}

\begin{document}
\maketitle
\begin{abstract}
SERRANT is a system and code for automatic classification of English grammatical errors that combines SErCl \citep{Choshen2020ClassifyingSE} and ERRANT \citep{errant}. SERRANT uses ERRANT's annotations when they are informative and those provided by SErCl otherwise. 
\end{abstract}

\section{Inroduction}
In grammatical error correction (GEC), an erroneous part of a sentence and its correction is called an \textit{edit}. It is often useful to categorize edits into types, e.g., to improve results \cite{JunczysDowmunt2018ApproachingNG,kantor2019learning} and for system evaluation \citep{choshen2018automatic, bryant2019bea}. A set of edit types is called a \textit{taxonomy}. In current annotations, taxonomies always differ between datasets of different languages \citep{rozovskaya2019grammar, Lee2016OverviewON} and mostly differ between datasets of the same language \citep{dahlmeier2013building,berzak2016universal}. For English, there are two automatic edit type classifiers, ERRANT and SErCl following two different taxonomies. These classifiers can be applied to any dataset containing corrections\footnote{Edits can be extracted automatically as well, so a sentence and its correction suffice.} and are thus the instrument of choice whenever more than one dataset is used. We present an attempt to combine the two taxonomies and the two classifiers into one system, which present outputs in a unified way.\footnote{Code is found in \url{https://github.com/matanel-oren/serrant}}

ERRANT \citep{errant} is a rule-based classifier of errors for English. ERRANT types include general categories like Spelling, Morphology and more specific categories reflecting the dominant POS in the edit (e.g. \type{Adverb} or \type{Adjective}). These categories may be further refined through sub-categories such as Inflection (e.g., \type{VERB:Infl}). ERRANT was used as the official taxonomy of the BEA shared task \citep{bryant2019bea} and provides a precise set of rules to map edits to types. 

A recent study presented SErCl \citep{Choshen2020ClassifyingSE}, a cross-lingual taxonomy of syntactic errors. SErCl defines an error type as the concatenation of morphosyntactic features of the text fragment before and after the change.  Formally, given a span from a learner sentence $l$ and its correction $c$, and given a Universal Dependencies \citep{nivre2016universal} annotation for these spans, such as UPOS tags, dependency-relation labels, or morphological-feature specifications for the heads of the spans' subtrees, which we denote as $UD_l$ and $UD_c$ respectively, the type of the error is $UD_l\shortrightarrow UD_c$. Thus, the type for an edit where a noun becomes a verb would be \typetwo{Noun}{Verb}, and when a noun changes its number from plural to singular, the type is \typetwo{Noun:singular}{Noun:plural}. If $UD_l$ is identical to $UD_c$, we denote this type with $UD_l$ for brevity, and if the span was deleted or added, the type is $UD_l$\ra\type{None} or \type{None}\ra $UD_c$ respectively.

\citet{Choshen2020ClassifyingSE} showed that some of ERRANT's categories are not informative or not consistent, while others are. They also note that ERRANT's POS-based types sometimes include cases where the POS changes. When POS is changed upon correction ERRANT is not well defined, it may assume the dominant type is the source or the correction, and quite often no type would be attached to the edit known as \type{other} type. ERRANT does have the benefit of being more human-readable when it is accurate, hence, SERRANT's default behaviour is chosen to be ERRANT types.

The main aim when constructing SERRANT's rules was informativeness. Thus, we preserve ERRANT's subclassifications allowing users to group together similar classes or ignore them altogether.

\section{Combining the classifiers}
In the default case, SERRANT returns ERRANT's edit types.\footnote{Following ERRANT's conventions, we keep the R, M, U initials meaning \textit{replacement}, \textit{missing}, and \textit{unnecessary} respectively. For example, a deleted verb, \typetwo{Verb}{None} in SErCl's terms, would be tagged U:Verb.} The special cases are the following:

\begin{enumerate}
    \item ERRANT's \type{other} category signifies failure to find informative type. Hence,  we rely on SErCl types in this case. We do keep the \type{other} category for unreliable cases, which we define as edits involving Intj, Num, Sym, X, and Punct POS tags. We also find that proper noun (Propn) is generally unreliable since the current parser uses it as a fallback for various erroneously spelled words. We do keep the \typetwo{Propn}{Propn} type because repetition reduces the risk of parser errors.
    \item ERRANT's morphology error type (\type{MORPH}) comprises a multitude of phenomena. We replace it with SErCl's types. We also expose the information originally captured by the \type{MORPH} type in a different way. When the lemmas do not match between the source and the target, we add a sub-category suffix \enquote{WC}. This indicates that while the POS did not change, the main error is in the choice of word and not in morphosyntactic features (e.g., \textit{consume}\ra \textit{eat} would be \type{VERB:WC} but \textit{eat}\ra \textit{ate} would not). We ignored the problematic cases mentioned in connection with the \type{other} ERRANT type but kept cases of \typetwo{Adj}{Propn} or \typetwo{Propn}{Adj}, such as \textit{China}\ra \textit{Chinese}.
    \item We added a suffix \enquote{MW}, which corresponds to a multi-word change in either the source or the correction. MW is only added when the multiword is not of an already named type such as \type{VERB:Tense}.
    \item ERRANT's orthography (\type{Orth}) type is generally correct, for example when reflecting a missing whitespace. However, it does contain cases where a proper noun should have been capitalized. While this is an orthographic error, unlike most such errors, it sometimes changes morphosyntax and/or the meaning. Therefore, if the word was not the first in the sentence, and was changed into a proper noun (Propn), SERRANT returns SErCl's annotation \typetwo{X}{propn} (e.g., \enquote{He founded \textit{apple}}~\ra\, \enquote{He founded \textit{Apple}}).
    \item The ERRANT \type{VERB} type reflects both \type{AUX} and \type{VERB} edits. We follow SErCl and mark them as \type{AUX} where needed.
    \item When a noun is changed into a verb, ERRANT marks this as \type{VERB:Form}. As this is not a change in the verb form, we denote this with \typetwo{Noun}{Verb}. (e.g., \textit{trap}\ra \textit{trapped}).
    \item Cases where a pronoun becomes a determiner or the other way around (e.g.\ \textit{these}\ra \textit{their}) are included in the ERRANT types \type{PRON} and determiner \type{DET}. We replace these annotations with more informative types \typetwo{Pron}{Det} and \typetwo{Det}{Pron}.
    \item ERRANT lumps tense, aspect, and mood together under \type{VERB:Tense}. When both the original and the corrected wordform have the lemmas \textit{be} or \textit{have} or the wordform is "will" we name it \type{VERB:Tense}. When both words are modal verbs (\textit{can}, \textit{could}, \textit{may}, \textit{might}, \textit{shall}, \textit{should}, \textit{will}, \textit{would}, \textit{must}), we add the suffix \enquote{Modal}. Otherwise we return the annotation provided by SErCl.
    
\end{enumerate}

\section{Examples}
In this section, we give some examples of the annotations returned by the model. The errors and corrections themselves are made up. The model follows the m2 format, but, for convenience, we provide a more visual format. More examples are provided with the code. In the examples \ra represents a correction of a given word (no multiword errors in the examples for simplicity), in brackets are the type SERRANT would give the error.

\begin{itemize}
    \item I werk\ra work (\type{R:Spell}) for pen\ra Pen (\typetwo{R:Noun}{Propn})
    \item gilly\ra Gilly (\type{R:Orth}) is imagination \ra imagining (\typetwo{R:Noun}{Verb})
    \item I drive\ra ride (\type{R:Verb:WC}) the\ra $\emptyset$ (\type{U:Det}) my bicycle.
    \item I should\ra shall (\type{R:Modal}) do as as I must.
\end{itemize}

We also add some examples from level A learners of the W\&I corpus \citep{bryant-etal-2019-bea}:

\begin{itemize}
    \item In addition to it\ra that (\typetwo{R:PRON}{DET}), we can also take a comfortable short nap on the back seat and wake up fresh.
    \item My family think that my cook\ra cooking (\type{R:MORPH:NOUN}) is amazing.
    \item It is great\ra very (\typetwo{R:ADV}{ADJ}) fun\ra funny (\typetwo{R:ADJ}{NOUN}).
    \item How are you? I'm writing to inform\ra give (\type{R:VERB:WC}) you that\ra $\emptyset$ (\type{U:PREP}) some advice on travelling and working in my country.
\end{itemize}

\bibliography{acl2020}
\bibliographystyle{acl_natbib}
\end{document}